\documentclass[11pt,a4paper]{article}
\usepackage[hyperref]{acl2017}
\usepackage{times}
\usepackage{latexsym}
\usepackage{url}
\usepackage{amsmath}
\usepackage{amssymb}
\usepackage{amsfonts}
\usepackage{graphicx}
\usepackage[labelfont=bf,skip=5pt]{caption}
\usepackage{subcaption}
\usepackage{array}
\usepackage{tabu}
\usepackage{makecell}
\usepackage{subcaption}
\usepackage{paralist}
\usepackage{cases}
\usepackage{diagbox}
\usepackage{enumitem}
\usepackage{soul}
\usepackage{multirow}
\usepackage{verbatim}
\usepackage{tabulary}
\usepackage{booktabs}
\usepackage[mathscr]{euscript}

\usepackage{color}
\usepackage{bm}
\definecolor{orange}{rgb}{1,0.5,0}
\definecolor{mdgreen}{rgb}{0.05,0.6,0.05}
\definecolor{mdblue}{rgb}{0,0,0.7}
\definecolor{dkblue}{rgb}{0,0,0.5}
\definecolor{dkgray}{rgb}{0.3,0.3,0.3}
\definecolor{slate}{rgb}{0.25,0.25,0.4}
\definecolor{gray}{rgb}{0.5,0.5,0.5}
\definecolor{ltgray}{rgb}{0.7,0.7,0.7}
\definecolor{purple}{rgb}{0.7,0,1.0}
\definecolor{lavender}{rgb}{0.65,0.55,1.0}

\newcommand{\ensuretext}[1]{#1}
\newcommand{\marker}[2]{\ensuremath{^{\textsc{#1}}_{\textsc{#2}}}}
\newcommand{\arkcomment}[3]{\ensuretext{\textcolor{#3}{[#1 #2]}}}
 \renewcommand{\arkcomment}[3]{}  

\newcommand{\stcomment}[1]{\arkcomment{\marker{S}{T}}{#1}{orange}}

\newcommand{\ra}[1]{\renewcommand{\arraystretch}{#1}}
\newcommand{\term}[1]{\textbf{#1}} 
\newcommand{\tensor}[1]{\mathbf{#1}}

\newcommand{\ledge}[3]{#1 \overset{#3}{\rightarrow} #2}
\newcommand{\edge}[2]{#1 {\rightarrow} #2}
\newcommand{\predicate}[1]{#1 {\rightarrow} \mathbf{\cdot}}
\newcommand{\bilstm}{\tensor{h}}
\newcommand{\fw}[1]{\overrightarrow{\bilstm}_{#1}}
\newcommand{\bw}[1]{\overleftarrow{\bilstm}_{#1}}
\newcommand{\shared}{\widetilde{\bilstm}}
\newcommand{\R}{\mathbb{R}}
\newcommand{\localscore}{s}
\newcommand{\globalscore}{S}
\newcommand{\factorpart}{p}
\newcommand{\adcubed}{$\text{AD}^3$}
\newcommand{\pred}{predicate}
\newcommand{\unlabeledarc}{unlabeled arc}
\newcommand{\labeledarc}{labeled arc}
 
\newcommand{\interalia}[1]{\citep[\emph{inter alia}]{#1}}

\newcommand{\argmax}[1]{\underset{#1}{\operatorname{arg}\,\operatorname{max}}\;}

\newcommand{\rulesep}{\unskip\ \vrule\ }
\setul{1pt}{.4pt}

\DeclareCaptionFont{tiny}{\tiny}
\aclfinalcopy 
\def\aclpaperid{596} 

\setitemize{noitemsep,topsep=10pt,parsep=0pt,partopsep=0pt}
\setenumerate{noitemsep,topsep=10pt,parsep=0pt,partopsep=0pt}

\title{Deep Multitask Learning for Semantic Dependency Parsing}

\author{
  Hao Peng$^\ast$ \quad
  Sam Thomson$^\dagger$ \quad
  Noah A. Smith$^\ast$ \\
  $^\ast$Paul G. Allen School of Computer Science \& Engineering,
  University of Washington, Seattle, WA, USA \\
   $^\dagger$School of Computer Science, Carnegie Mellon University, Pittsburgh, PA, USA\\
   {\tt \{hapeng,nasmith\}@cs.washington.edu, sthomson@cs.cmu.edu}
}
   
\date{}

\begin{document}

\maketitle

\begin{abstract}
We present a deep neural architecture that parses sentences into
three semantic dependency graph formalisms.
By using efficient, nearly arc-factored inference and a bidirectional-LSTM
composed with a multi-layer perceptron, 
our base system is able to significantly improve the state of the art
for semantic dependency parsing, without using hand-engineered
features or syntax.
We then explore two multitask learning approaches---one that shares
parameters across formalisms, and one that uses higher-order structures to predict
the graphs jointly.
We find that both approaches improve performance across formalisms on
average, achieving a new state of the art.
Our code is open-source and available at
\url{https://github.com/Noahs-ARK/NeurboParser}.
\end{abstract}

\section{Introduction}
\label{sec:intro}

\begin{figure}
  \centering
  \begin{subfigure}[b]{\columnwidth}
    \label{subfig:dm}
    \includegraphics[width=\columnwidth]{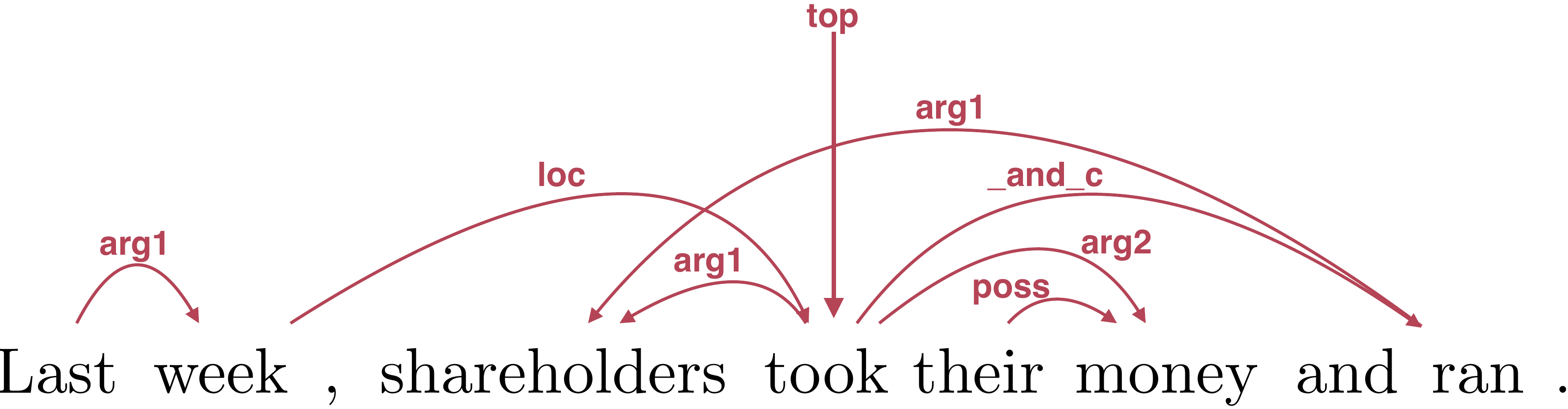}
    \caption{DM}
  \end{subfigure}
  \begin{subfigure}[b]{\columnwidth}
    \label{subfig:pas}
    \includegraphics[width=\columnwidth]{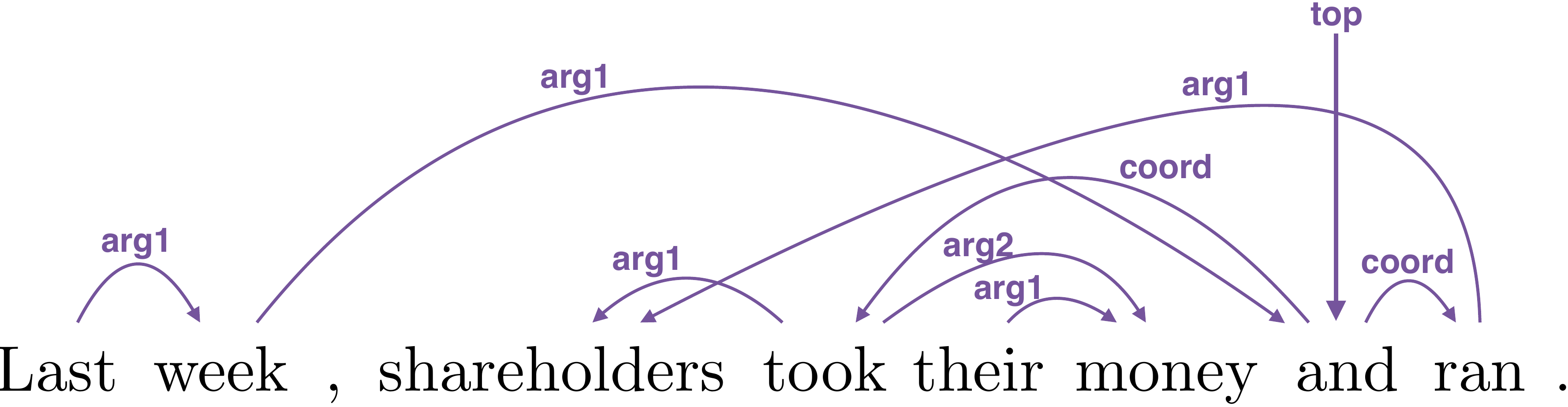}
    \caption{PAS}
  \end{subfigure}
  \begin{subfigure}[b]{\columnwidth}
    \label{subfig:psd}
    \includegraphics[width=\columnwidth]{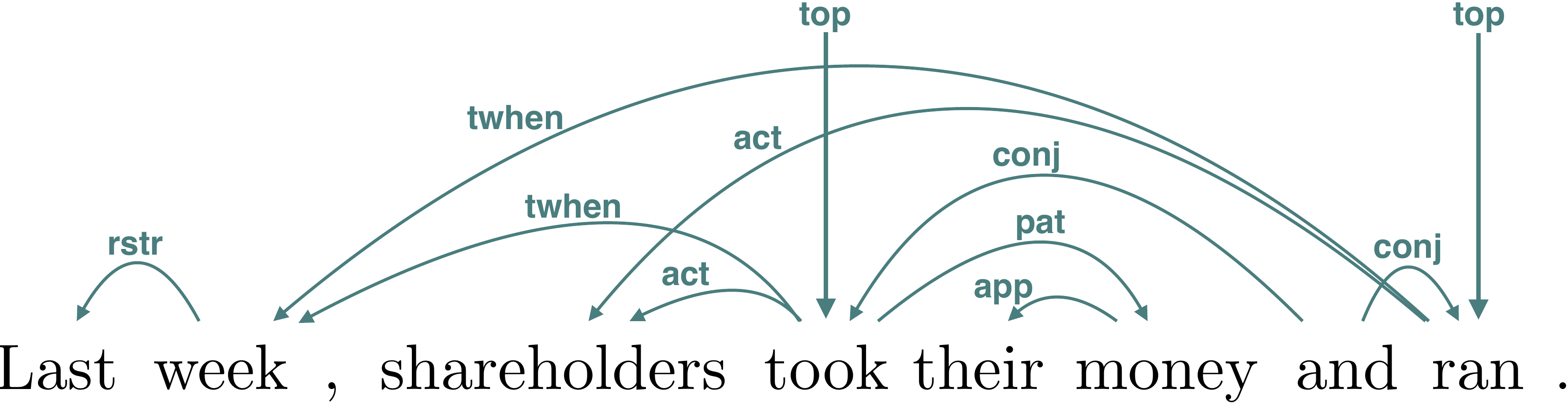}
    \caption{PSD}
  \end{subfigure}
  \caption{An example sentence annotated with the three semantic formalisms of
  the broad-coverage semantic dependency parsing shared tasks.}
  \vspace{-.25cm}
  \label{fig:formalisms}
\end{figure}

\stcomment{
Motivate more why semantic parsing is worth doing.
Information extraction, QA, \ldots? (find relevant cites)
}
\stcomment{
Maybe put our raw numbers earlier, because they are quite good compared to
numbers people expect from other semantic parsing tasks.
This may make the difference between semantic parsing being ``useful'' and vs.
not.
}

Labeled directed graphs are a natural and flexible representation for semantics
\interalia{copestake_minimal_2005,baker_semeval_2007,surdeanu_conll_2008,banarescu_abstract_2013}.
Their generality over trees, for instance, allows them to represent
relational semantics while handling phenomena like coreference and
coordination.
Even syntactic formalisms are moving toward graphs
\citep{Marneffe2014UniversalSD}.
However, full semantic graphs can be expensive to annotate,
and efforts are fragmented across competing semantic theories, leading to
a limited number of annotations in any one formalism.
This makes learning to parse more difficult, especially for
powerful but data-hungry machine learning techniques like neural networks.

In this work, we hypothesize that the overlap among theories and
their corresponding representations can be exploited using multitask
learning \citep{caruana_multitask_1997}, allowing us to learn from more data.
We use the 2015 SemEval shared task on Broad-Coverage Semantic Dependency
Parsing \citep[\term{SDP};][]{oepen2015sdp} as our testbed.
The shared task provides an English-language corpus with parallel annotations
for three semantic graph representations, described in \S\ref{sec:task}.
Though the shared task was designed in part to encourage comparison between the
formalisms, we are the first to treat SDP as a multitask learning problem.

As a strong baseline, we introduce a new system that parses each formalism
separately  (\S\ref{sec:mono_task}).
It uses a bidirectional-LSTM composed with a multi-layer perceptron to score
arcs and predicates, and has efficient, nearly arc-factored inference.
Experiments show it significantly improves on state-of-the-art methods
(\S\ref{subsec:mono:experiment}).



We then present two multitask extensions 
(\S\ref{subsec:multi:sharing} and \S\ref{subsec:multi:ct}), with a
parameterization and factorization that implicitly models the relationship between multiple formalisms.
Experiments show that both techniques improve over our basic model, with an
additional (but smaller) improvement when they are combined
(\S\ref{subsec:multi:experiment}).
Our analysis shows that the improvement in unlabeled ${F}_1$ is greater for the
two formalisms that are more structurally similar, and suggests directions for future work.
Finally, we survey related work (\S\ref{sec:related}), and
summarize our contributions and
findings (\S\ref{sec:conclusion}).

\section{Broad-Coverage Semantic Dependency Parsing (SDP)}
\label{sec:task}

\begin{table}[tb]
\small
		\begin{tabulary}{\columnwidth}{@{}l RR c RR c RR}
			\toprule
			& \multicolumn{2}{c}{\bf{DM}}
			& \phantom{ }
			& \multicolumn{2}{c}{\bf{PAS}}
			& \phantom{ }
			& \multicolumn{2}{c}{\bf{PSD}}\\
			\cmidrule{2-3}
			\cmidrule{5-6}
			\cmidrule{8-9}
			& id & ood
			& & id & ood
			& & id & ood  \\
			\midrule
		 	\# labels & 59 & 47 && 42 & 41 && 91 & 74\\
			 \% trees & 2.3 & 9.7 && 1.2 & 2.4 && 42.2 & 51.4 \\
			 \% projective & 2.9 & 8.8 && 1.6 & 3.5 && 41.9 & 54.4 \\
			\bottomrule
		\end{tabulary}
	\caption{Graph statistics for in-domain (WSJ, ``id'') and
          out-of-domain (Brown corpus, ``ood'') data. Numbers taken from \citet{oepen2015sdp}.} 
	\label{tab:data} 
	\vspace{-.25cm}
\end{table}

\stcomment{
MRS argues that graphs are a more natural representation for
semantics than, say, recursive tree structures, because they underspecify the
order of composition in the right way;
i.e. derivation trees have spurious ambiguity w.r.t. the semantics.
For example, the semantics of intersective adjectives are commutative
and associative.
So the fact that ``fierce black cat'' has a different branching structure
than ``gato negro y feroz'' obscures the fact that they have the same
meaning \citep{copestake_minimal_2005}.
}
\stcomment{Could also compare to lambda-calculus logical forms.}
\stcomment{I think I'm just going to leave these points out.}

First defined in a SemEval 2014 shared task \citep{oepen2014sdp}, and then
extended by \citet{oepen2015sdp}, the broad-coverage semantic depency parsing (\term{SDP})
task is centered around three semantic formalisms whose annotations have been
converted into bilexical dependencies.
See Figure~\ref{fig:formalisms} for an example.
The formalisms come from varied linguistic traditions, but all three
aim to capture predicate-argument relations between content-bearing
words in a sentence.

While at first glance similar to syntactic dependencies, semantic
dependencies have distinct goals and characteristics, more akin to semantic
role labeling \citep[SRL;][]{gildea2002srl} or the abstract meaning
representation \citep[AMR;][]{banarescu_abstract_2013}.
They abstract
over different syntactic realizations of the same or similar meaning
(e.g., \textit{``She gave me the ball.''} vs. \textit{``She gave the ball to
me.''}).
Conversely, they attempt to distinguish between different senses even when
realized in similar syntactic forms (e.g., \textit{``I baked in the kitchen.''}
vs. \textit{``I baked in the sun.''}).

Structurally, they are labeled directed graphs whose vertices are tokens in the
sentence.
This is in contrast to AMR whose vertices are abstract concepts, with no
explicit alignment to tokens, which makes parsing more
difficult~\citep{flanigan2014amr}.
Their arc labels encode broadly-applicable semantic relations rather
than being tailored to any specific downstream application or ontology.\footnote{%
This may make another disambiguation step necessary to use these
representations in a downstream task, but there is evidence that modeling
semantic composition separately from grounding in any
ontology is an effective way to achieve broad
coverage~\citep{Kwiatkowski2013ScalingSP}.
}
They are not necessarily trees, because a token may be an argument of more than
one predicate (e.g.,  in \textit{``John wants to eat,''} John is both the wanter and the would-be eater).
Their analyses may optionally leave out non--content-bearing tokens, such as
punctuation or the infinitival \textit{``to,''} or prepositions that simply
mark the type of relation holding between other words.
But when restricted to content-bearing tokens (including adjectives, adverbs,
etc.), the subgraph is connected.
In this sense, SDP provides a \emph{whole-sentence} analysis.
This is in contrast to PropBank-style SRL, which gives an analysis of
only verbal and nominal predicates \citep{palmer2005proposition}.
Semantic dependency graphs also tend to have higher levels of nonprojectivity
than syntactic trees \citep{oepen2014sdp}.
Sentences with graphs containing cycles have been removed from the dataset by
the organizers, so all remaining graphs are directed acyclic graphs.
Table \ref{tab:data} summarizes some of the dataset's high-level statistics.

\paragraph{Formalisms.}
Following the SemEval shared tasks, we consider three formalisms.
The \term{DM} (DELPH-IN MRS) representation comes from DeepBank \citep{flickinger_deepbank_2012}, which
are manually-corrected parses from the LinGO English Resource Grammar
\citep{copestake_erg_2000}.
LinGO is a head-driven phrase structure grammar
\citep[HPSG;][]{pollard_hpsg_94} with minimal recursion semantics \citep{copestake_minimal_2005}.
The \term{PAS} (Predicate-Argument Structures) representation is extracted from the Enju Treebank, which
consists of automatic parses from the Enju HPSG parser \citep{miyao_linguistic_2006}.
PAS annotations are also available for the Penn Chinese Treebank
\citep{xue2005ctb}.
The \term{PSD} (Prague Semantic Dependencies) representation is extracted from the tectogrammatical layer
of the Prague Czech-English Dependency Treebank \citep{hajic2012psd}.
PSD annotations are also available for a Czech translation of the WSJ Corpus.
In this work, we train and evaluate only on English annotations.

Of the three, PAS follows syntax most closely, and prior work has found it the easiest to predict.
PSD has the largest set of labels, and parsers have significantly lower performance on it
\citep{oepen2015sdp}.


\section{Single-Task SDP}
\label{sec:mono_task}

Here we introduce our basic model, in which training and prediction for each
formalism is kept completely separate.
We also lay out basic notation, which will be reused for our
multitask extensions.

\subsection{Problem Formulation}
\label{subsec:mono:formulation}

\begin{figure}
 \small
  \centering
  \begin{subfigure}[b]{.3\columnwidth}
    \includegraphics[width=\columnwidth]{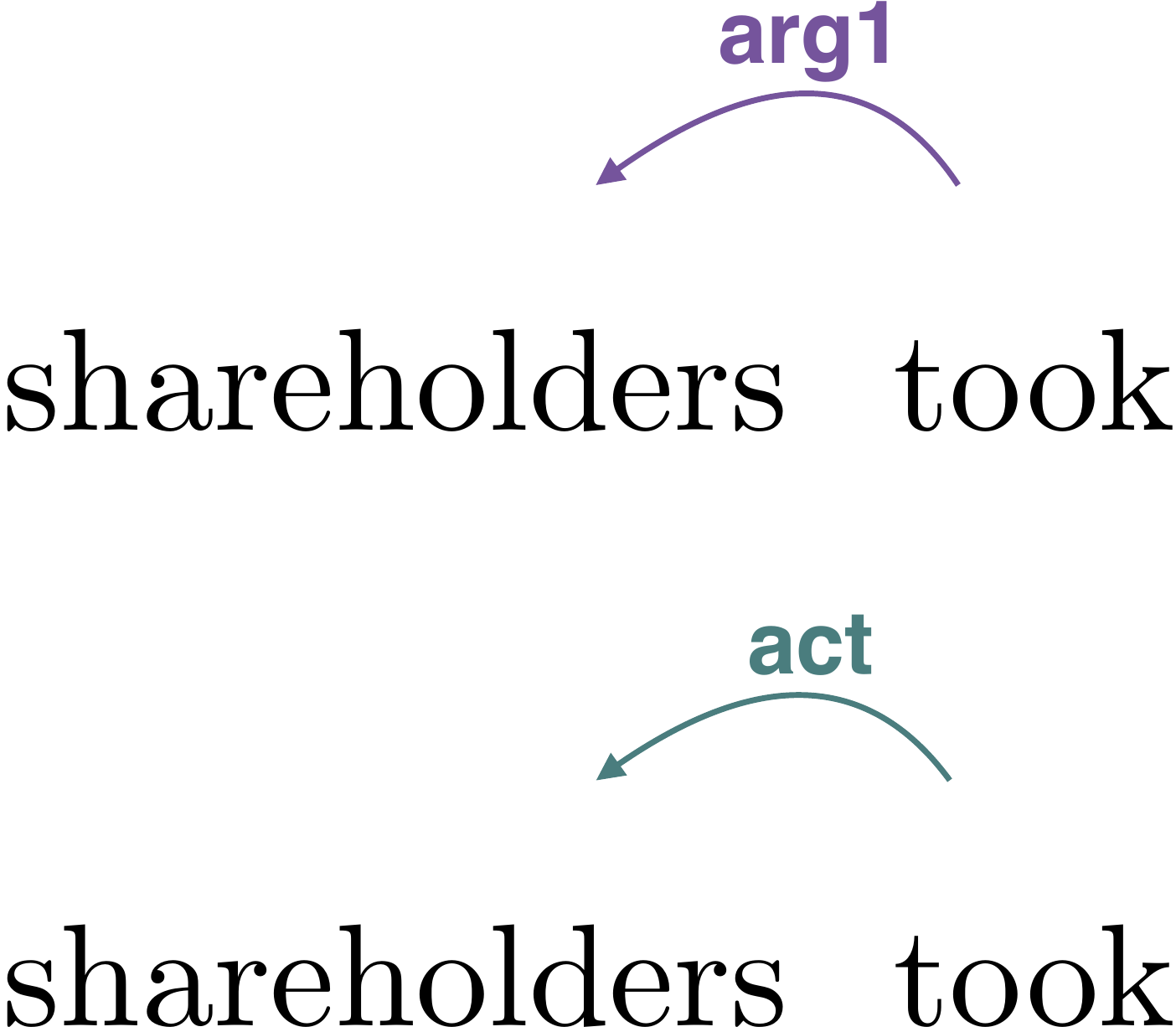}
    \caption{First-order.}
    \label{subfig:1st_order_factor_arg2}
  \end{subfigure}
  \rulesep
  \begin{subfigure}[b]{.3\columnwidth}
    \includegraphics[width=\columnwidth]{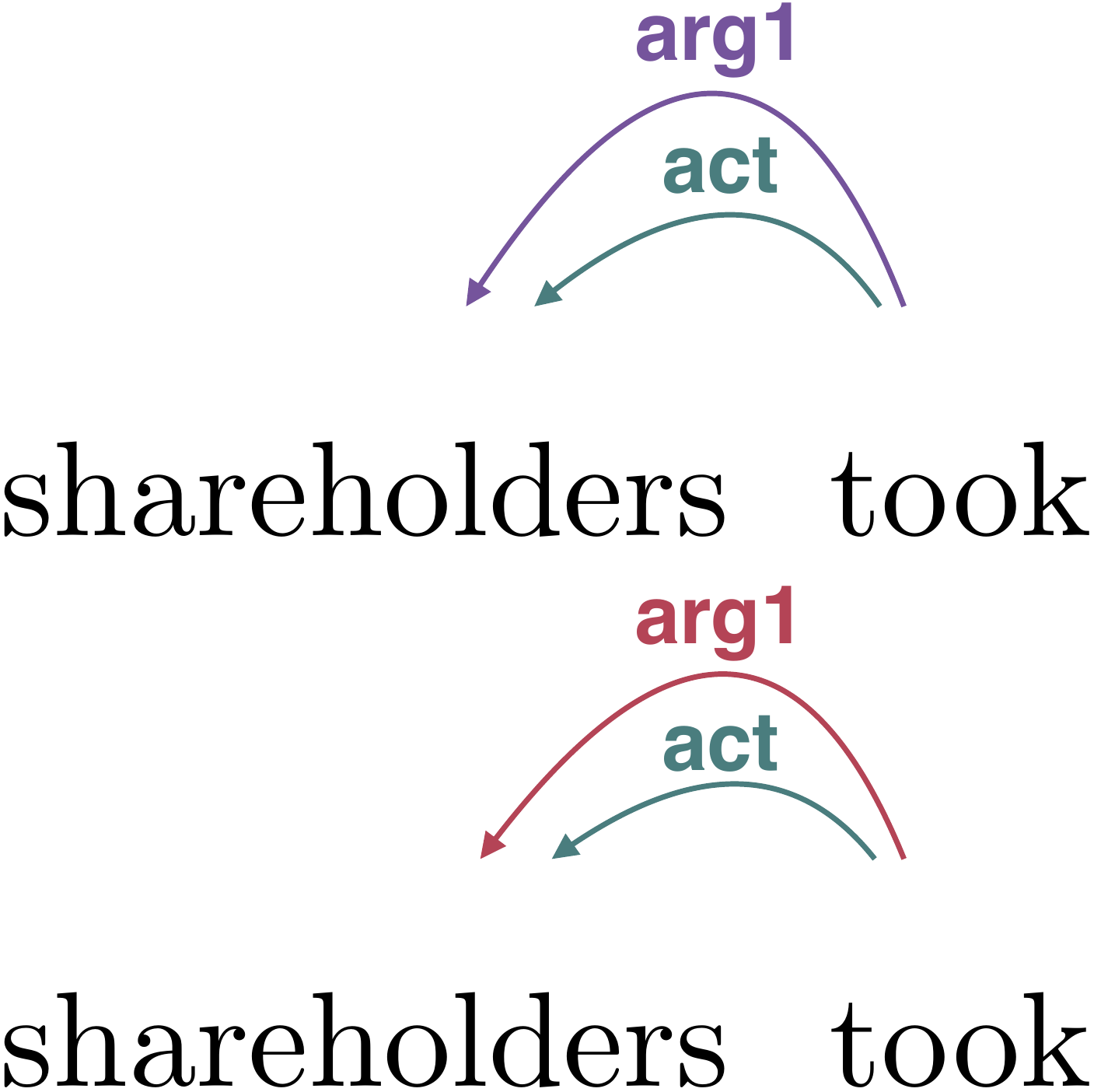}
    \caption{Second-order.}
    \label{subfig:2nd_order_factor}
  \end{subfigure}
  \rulesep
  \begin{subfigure}[b]{.3\columnwidth}
    \includegraphics[width=\columnwidth]{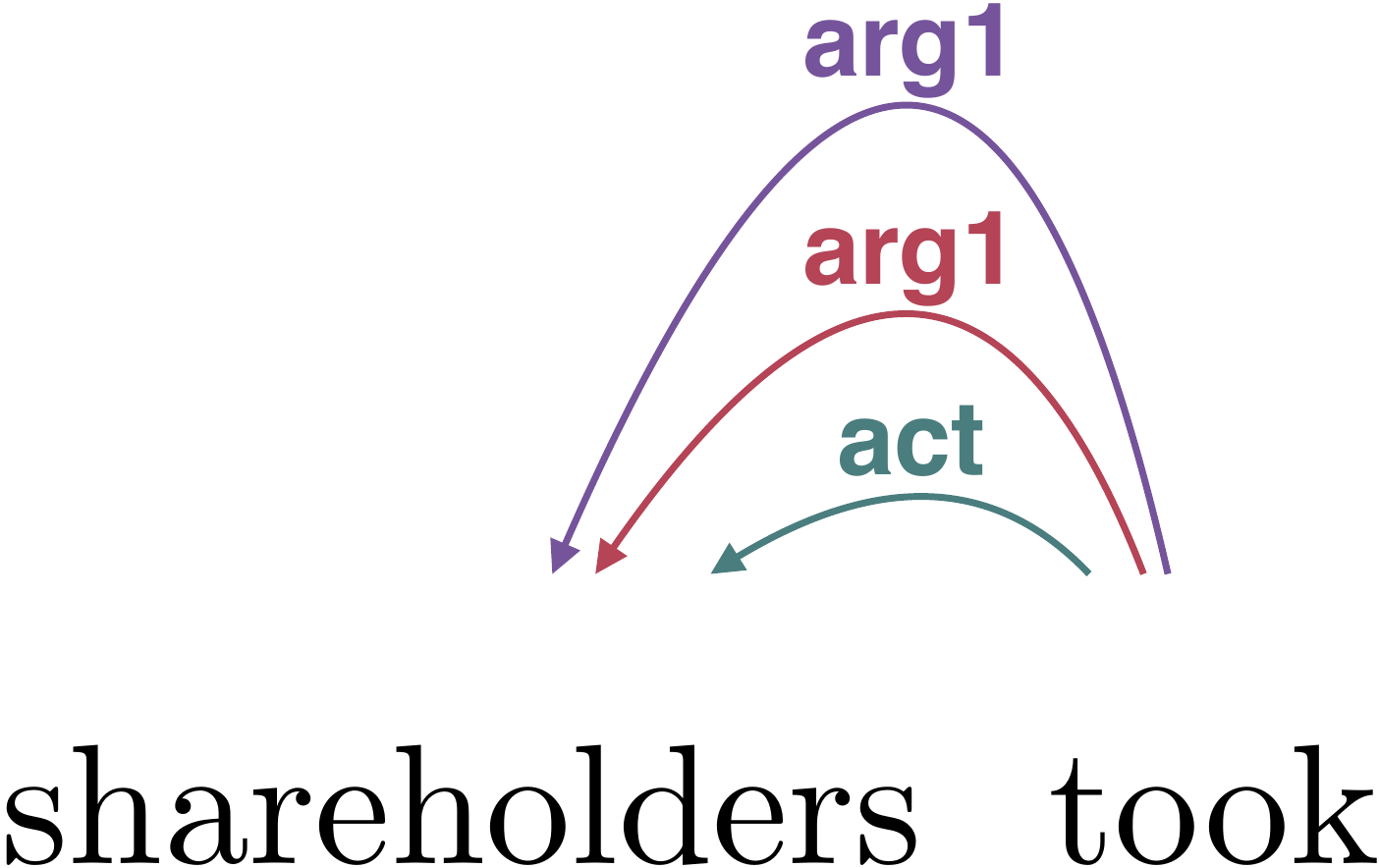}
    \caption{Third-order.}
    \label{subfig:3rd_order_factor}
  \end{subfigure}
  \caption{Examples of local structures. We refer to the number
  of arcs that a structure contains as its \term{order}.}
  \vspace{-.4cm}
  \label{fig:substructures}
\end{figure}

The output of semantic dependency parsing is a labeled directed graph
(see Figure~\ref{fig:formalisms}).
Each arc has a label from a predefined set $\mathcal{L}$\stcomment{$^{(t)}$?},
indicating the semantic relation of the child to the head. 
Given input sentence $x$, let $\mathcal{Y}(x)$ be the set of possible semantic
graphs over $x$.
The graph we seek maximizes a score function
$\globalscore$:
\begin{equation}
\hat{y} = \argmax{y \in\mathcal{Y}(x)} \globalscore(x, y), 
\label{eq:score}
\end{equation}
We decompose $\globalscore$ into a sum of local scores 
$s$ for \textbf{local structures} (or ``parts'') $p$ in the graph:
\begin{equation}
\globalscore(x, y) = \sum_{\factorpart \in y} {
\localscore(\factorpart).
}
\label{eq:score_decompose}
\end{equation}
For notational simplicity, we omit the dependence of $\localscore$ on
$x$.
See Figure~\ref{subfig:1st_order_factor_arg2} for examples of local
structures.
$\localscore$ is a parameterized function, whose parameters (denoted
$\Theta$ and suppressed here for clarity) will be learned from
the training data (\S\ref{subsec:learning}).
Since we search over every possible labeled graph (i.e., considering
each labeled arc for each pair of words), our approach can be considered a
\term{graph-based} (or \term{all-pairs}) method.
The models presented in this work all share this common
graph-based approach, differing only in the set of
structures they score and in the parameterization of the
scoring function $\localscore$. This approach also underlies 
state-of-the-art approaches to SDP \cite{martins2014sdp}. 

\subsection{Basic Model}
\label{subsec:mono:basic}

Our basic model is inspired by recent successes in
neural arc-factored graph-based dependency parsing
\citep{kiperwasser2016simple,dozat2016deep,kuncoro-16}.
It borrows heavily from the neural arc-scoring architectures in those works, but
decodes with a different algorithm under slightly different constraints.

\subsubsection{Basic Structures}

Our basic model factors over three types of structures ($\factorpart$ in Equation~\ref{eq:score_decompose}):
\begin{compactitem}
\item\pred, indicating a predicate word, denoted $\predicate{i}$;
\item\unlabeledarc, representing the existence of an arc from a
predicate to an argument, denoted $\edge{i}{j}$;
\item\labeledarc, an arc labeled with a semantic role, denoted $\ledge{i}{j}{\ell}$.
\end{compactitem}
Here $i$ and $j$ are word indices in a given sentence, and $\ell$ indicates the arc label.
This list corresponds to the most basic structures used by
\citet{martins2014sdp}.  Selecting an output $y$ corresponds precisely
to selecting which instantiations of these structures are included.

To ensure the internal consistency of predictions, the following constraints
are enforced during decoding:
\begin{compactitem}
\item $\predicate{i}$ if and only if there exists at least one $j$ such that
$\edge{i}{j}$;
\item If $\edge{i}{j}$, then there must be exactly one label $\ell$ such that
$\ledge{i}{j}{\ell}$.
Conversely, if not $\edge{i}{j}$, then there must not exist any
$\ledge{i}{j}{\ell}$;
\end{compactitem}
We also enforce a \term{determinism} constraint \citep{flanigan2014amr}:
certain labels must not appear on more than one arc emanating from the same token.
The set of deterministic labels is decided based on their appearance in the training set.
Notably, we do not enforce that the predicted graph is connected or spanning.
If not for the \pred\ and determinism constraints, our model would be
\term{arc-factored}, and decoding could be done for each $i, j$ pair
independently.
Our structures do overlap though, and we employ \adcubed\ \citep{martins2011dual}
to find the highest-scoring internally consistent semantic graph.
\adcubed\ is an approximate discrete optimization algorithm based on dual
decomposition.
It can be used to decode factor graphs over discrete variables when
scored structures
overlap, as is the case here.


\subsubsection{Basic Scoring}
\label{subsec:mono:scoring}

Similarly to \citet{kiperwasser2016simple}, 
our model
learns representations of tokens in a sentence using a bi-directional LSTM (BiLSTM).
Each different type of structure (\pred, \unlabeledarc, \labeledarc) then shares
these same BiLSTM representations, feeding them into
a multilayer perceptron (MLP) which is specific to the structure type.
We present the architecture slightly differently from prior work, to make the
transition to the multitask scenario (\S\ref{sec:multi_task}) smoother.
In our presentation, we separate the model into a function $\bm{\phi}$ that
represents the input (corresponding to the BiLSTM and the initial layers of the
MLPs), and a function $\bm{\psi}$ that represents the output (corresponding to
the final layers of the MLPs), with the scores given by their inner
product.\footnote{For clarity, we present single-layer BiLSTMs and MLPs, 
  while in practice we use two layers for both.
}

%
\paragraph{Distributed input representations.}
Long short-term memory networks (LSTMs) are a variant of recurrent neural
networks (RNNs) designed to alleviate the vanishing gradient problem in
RNNs \citep{hochreiter_long_1997}.
A bi-directional LSTM (BiLSTM) runs over the sequence in both directions
\citep{shcuster1997bidirectional,graves2012supervised}.

Given an input sentence $x$ and its corresponding part-of-speech tag sequence,
each token is mapped to a concatenation of its word embedding vector and POS tag
vector.
Two LSTMs are then run in opposite directions over the input vector sequence,
outputting the concatenation of the two hidden vectors at each position
$i$:  $\bilstm_i = \bigl[\fw{i}; \bw{i}\bigr]$
(we omit $\bilstm_i$'s dependence on $x$ and its own parameters).
$\bilstm_i$ can be thought of as an encoder that contextualizes each token
conditioning on all of its context, without any Markov assumption.
$\bilstm$'s parameters are learned jointly with the rest of the model
(\S\ref{subsec:learning}); we refer the readers to \citet{cho2015natural} for
technical details. 

\begin{figure}
	\includegraphics[width=\columnwidth]{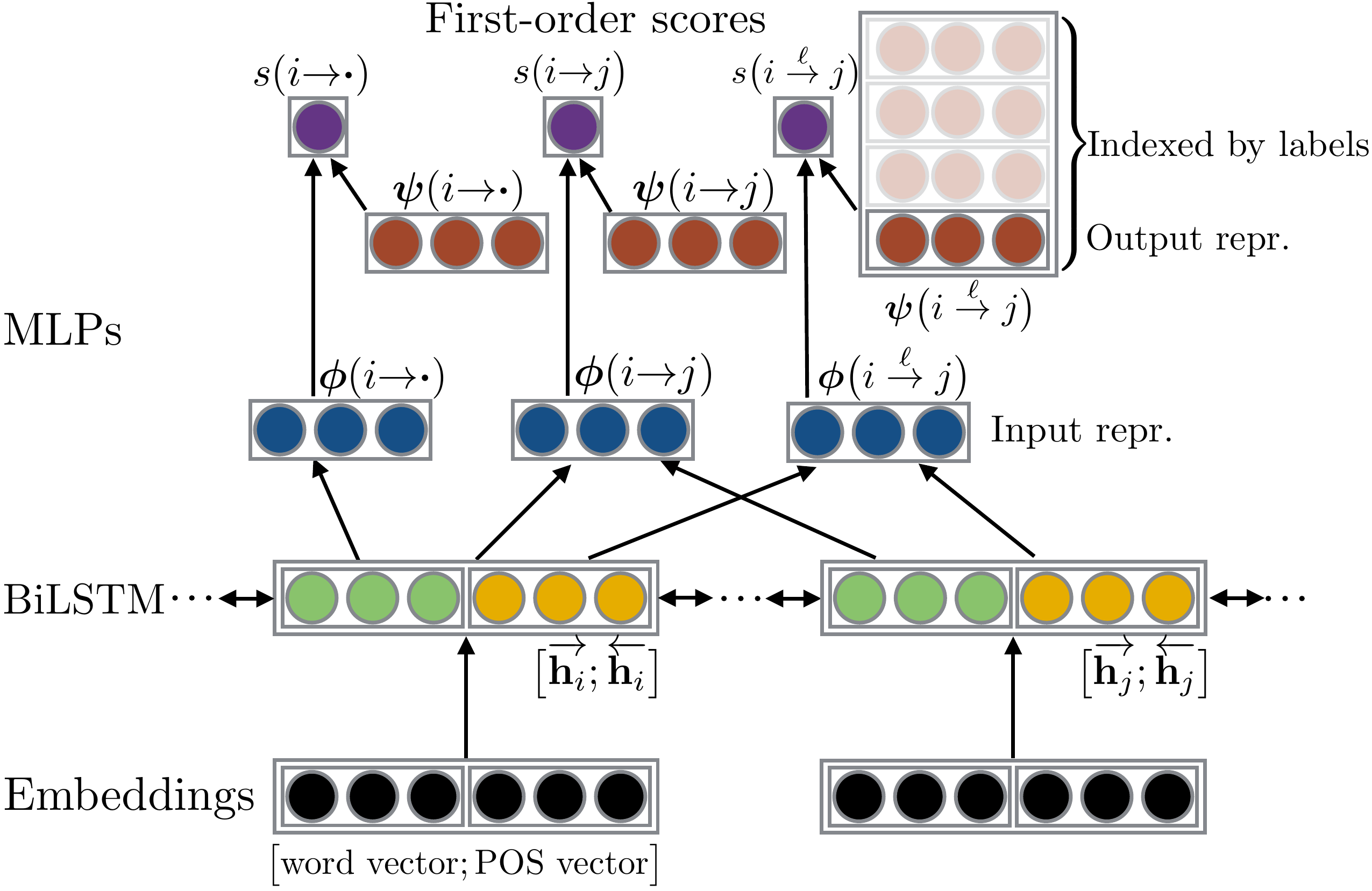}
	\caption{Illustration of the architecture of the basic model. $i$ and $j$ denote
          the indices of tokens in the given sentence. 
          The figure depicts single-layer BiLSTM and MLPs, while in practice we use two layers for both.
          }
	\label{fig:diagram_singletask}
\end{figure}

The input representation $\bm{\phi}$ of a \pred\ structure depends on the
representation of one word:
\begin{subequations}
\begin{align}
 \bm{\phi}(\predicate{i}) &=
 \tanh\bigl(\mathbf{C}_{\text{pred}} \bilstm_i +
 \mathbf{b}_\text{pred}\bigr). \label{eq:phi_pred}\\
\intertext{For \unlabeledarc\ and \labeledarc\ structures, it depends on both
the head and the modifier (but not the label, which is captured in the
distributed output representation):} \bm{\phi}(\edge{i}{j}) &=
 \tanh\bigl(\mathbf{C}_{\text{UA}} \bigl[\bilstm_i; \bilstm_j\bigr] +
 \mathbf{b}_\text{UA}\bigr), \label{eq:phi_ua}\\
 \bm{\phi}(\ledge{i}{j}{\ell}) &=
 \tanh\bigl(\mathbf{C}_{\text{LA}} \bigl[\bilstm_i; \bilstm_j\bigr] +
 \mathbf{b}_\text{LA}\bigr).  \label{eq:phi_la}
 \end{align}
\end{subequations}

\paragraph{Distributed output representations.}
NLP researchers have found that embedding discrete output labels into a low
dimensional real space is an effective way to capture commonalities among
them
\interalia{srikumar2014learning,hermann_semantic_2014,fitzgerald2015semantic}.
In neural language models \interalia{bengio_neural_2003,mnih_three_2007} the
weights of the output layer could also be regarded as an output embedding.

We associate each first-order structure $p$ with a $d$-dimensional real vector
$\bm{\psi}(p)$ which does not depend on particular words in $p$.
Predicates and unlabeled arcs are each mapped to a single vector:
\begin{subequations}
\begin{align}
\bm{\psi}(\predicate{i}) &= \bm{\psi}_{\text{pred}},  \label{eq:psi_pred}\\
\bm{\psi}(\edge{i}{j})  &= \bm{\psi}_{\text{UA}}, \label{eq:psi_ua}
\intertext{and each label gets a vector:}
\bm{\psi}(\ledge{i}{j}{\ell}) &= \bm{\psi}_{\text{LA}}(\ell).\label{eq:psi_la}
\end{align}
\end{subequations}

\paragraph{Scoring.}
Finally, we use an inner product to score first-order structures:
\begin{equation}
\label{eq:score_first_order}
s(p) = \bm{\phi}(p) \cdot \bm{\psi}(p).
\end{equation}
Figure~\ref{fig:diagram_singletask} illustrates our basic model's architecture.

\subsection{Learning}
\label{subsec:learning}
The parameters of the model are learned using a max-margin objective.
Informally, the goal is to learn parameters for the score function so that the
gold parse is scored over every incorrect parse with a margin proportional to
the cost of the incorrect parse.
More formally, let $\mathcal{D} = \bigl\{(x_i, y_i)\bigr\}_{i=1}^{N}$ be the training set consisting of $N$ pairs of sentence $x_i$ and its gold parse $y_i$. Training is then the following $\ell_2$-regularized empirical risk minimization problem:
\begin{align}
 \min_{\Theta}{\frac{\lambda}{2}\lVert \Theta \rVert^2 + \frac{1}{N}\sum_{i=1}^{N}L\bigl(x_i, y_i; \Theta\bigr)},
\end{align}
where $\Theta$ is all parameters in the model, and $L$ is the
structured hinge loss:
\begin{equation}
\begin{split}
L\bigl(x_i, y_i; \Theta\bigr) = \hspace{-.2cm}\max_{y \in \mathcal{Y}(x_i)} \bigl\{& \globalscore\bigl(x_i, y\bigr) + c\bigl(y, y_i\bigr)\bigr\}\\
 & - \globalscore\bigl(x_i, y_i\bigr).
\end{split}
\end{equation}
$c$ is a weighted Hamming distance that
trades off between precision and recall  \citep{taskar-04}.
Following \citet{martins2014sdp}, we encourage recall over precision by using
the costs 0.6 for false negative arc predictions and 0.4 for false positives.

\subsection{Experiments}
\label{subsec:mono:experiment}

We evaluate our basic model on the English dataset from SemEval 2015 Task 18
closed track.\footnote{\url{http://sdp.delph-in.net}}
We split as in previous work
\citep{almeida2015sdp,du2015sdp}, resulting in 33,964 training
sentences from \S 00--19 of the WSJ corpus, 1,692 development
sentences from \S 20, 1,410 sentences from \S 21 as
in-domain test data, and 1,849 sentences sampled from the Brown Corpus as out-of-domain test data.

The \emph{closed} track differs from the \emph{open} and \emph{gold} tracks in
that it does not allow access to any syntactic analyses.
In the open track, additional machine generated syntactic parses are
provided, while the gold-track gives access to various gold-standard syntactic analyses. 
Our model is evaluated with closed track data; it does not have access to any syntactic analyses during training or test. 

We refer the readers to \S\ref{sec:implementation}
for implementation details, including training procedures, hyperparameters, pruning techniques, etc..

\begin{table}[tb]
\center
		\begin{tabulary}{\textwidth}{@{}l  l   r r r  r@{}}
			\toprule

			& \textbf{Model}
			& \textbf{DM}
			& \textbf{PAS}
			& \textbf{PSD}
			& \textbf{Avg.}\\

			\midrule

			\multirow{3}{*}{id}
			& Du et al., 2015
			& 89.1 &  91.3
			& 75.7 & 86.3 \\

			& A\&M, 2015
			& 88.2 &  90.9
			& 76.4 & 86.0 \\

			& \textsc{basic}
			& \textbf{89.4} &  \ul{\textbf{92.2}}
			& \ul{\textbf{77.6}} & \ul{\textbf{87.4}} \\

			\midrule[\cmidrulewidth]

			\multirow{3}{*}{ood}
			& Du et al., 2015
			& 81.8 &  87.2
			& 73.3 & 81.7 \\
			
			& A\&M, 2015
			& 81.8 &  86.9
			& 74.8 & 82.0 \\
			
			& \textsc{basic}
			& \ul{\textbf{84.5}} &  \ul{\textbf{88.3}}
			& \ul{\textbf{75.3}} & \ul{\textbf{83.6}} \\
			
			\bottomrule
		\end{tabulary}
	\caption[Caption for LOF]{Labeled parsing
          performance ($F_1$ score) on both in-domain (id) and
          out-of-domain (ood) test data. The
          last column shows the micro-average over the three
          tasks. Bold font indicates best performance without syntax.
          Underlines indicate statistical significance
          with \citet{bonferroni36teoria} correction compared
          to the best baseline system.\protect\footnotemark}
          \vspace{-.25cm}
	\label{tab:mono_res_lf}
\end{table}
\footnotetext{Paired bootstrap, $p < 0.05$ after Bonferroni correction.}

\paragraph{Empirical results.}
As our model uses no explicit syntactic information, the most comparable models
to ours are two state-of-the-art closed track systems due to \citet{du2015sdp} and \citet{almeida2015sdp}.
\citet{du2015sdp} rely on graph-tree transformation techniques
proposed by \citet{du2014sdp}, and apply a voting ensemble to
well-studied tree-oriented parsers.
Closely related to ours is \citet{almeida2015sdp}, who used
rich, hand-engineered second-order features and \adcubed\ for inference.

Table~\ref{tab:mono_res_lf} compares our basic model to both baseline
systems (labeled $F_1$ score) on SemEval 2015 Task 18 test data.
Scores of those systems are repeated from the official evaluation results.
Our basic model significantly outperforms the best
published results with a 1.1\% absolute improvement on the in-domain test set
and 1.6\% on the out-of-domain test set.

\section{Multitask SDP}
\label{sec:multi_task}
We introduce two extensions to our single-task model,
both of which 
use  training data for all three formalisms to improve performance on each
formalism's parsing task.
We describe a first-order model, where representation functions are
enhanced by parameter sharing while inference is kept separate for each task (\S\ref{subsec:multi:sharing}).
We then introduce a model with cross-task higher-order structures that uses joint
inference \emph{across} different tasks (\S\ref{subsec:multi:ct}).
Both multitask models use
\adcubed\ for decoding, and are trained with the same margin-based objective,
as in our single-task model.

\subsection{Problem Formulation}
\label{{subsec:multi:formulation}}
We will use an
additional superscript $t \in \mathcal{T}$ to distinguish the three tasks
(e.g., $y^{(t)}$, $\bm{\phi}^{(t)}$), where $\mathcal{T} = \left\{\mathrm{DM}, \mathrm{PAS}, \mathrm{PSD}\right\}$.
Our task is now to predict three graphs
$\{y^{(t)}\}_{t\in\mathcal{T}}$ for a given input sentence $x$.
Multitask SDP can also be understood as parsing $x$ into a single unified  \emph{multigraph}
$y = \bigcup_{t\in\mathcal{T}}y^{(t)}$. 
Similarly to Equations~\ref{eq:score}--\ref{eq:score_decompose},
we decompose $y$'s score $\globalscore(x, y)$
into a sum of local scores for local structures in $y$,
and we seek a multigraph $\hat{y}$ that maximizes $\globalscore(x, y)$.

\subsection{Multitask SDP with Parameter Sharing}
\label{subsec:multi:sharing}
A common approach when using BiLSTMs for multitask learning is to share the
BiLSTM part of the model across tasks, while training specialized classifiers
for each task \citep{soggard2016deep}.
In this spirit, we let each task keep its own specialized MLPs, and explore
two variants of our model that share parameters at the BiLSTM level.

The first variant consists of a set of task-specific BiLSTM encoders 
as well as a common one that is shared across all tasks.
We denote it \textsc{freda}. 
\textsc{freda} uses a neural generalization of ``frustratingly easy'' domain
adaptation \citep{daume2007frustratingly,kim2016frustratingly}, where one augments
domain-specific features with a shared set of features to capture global patterns.
Formally, let $\{\bilstm^{(t)}\}_{t\in\mathcal{T}}$ denote the three task-specific
encoders.
We introduce another encoder $\shared$ that is shared across
all tasks.
Then a new set of input functions $\{\bm{\phi}^{(t)}\}_{t\in\mathcal{T}}$ can
be defined as in Equations~\ref{eq:phi_pred}--\ref{eq:phi_la}, for example:
\begin{equation}
\begin{split}
  \label{eq:phi_multitask_}
   \bm{\phi}^{(t)}(\ledge{i}{j}{\ell}) =
  \tanh\bigl(\mathbf{C}_{\text{LA}}^{(t)} &\bigl[  \bilstm^{(t)}_i ;
      \bilstm^{(t)}_j ; \\
      &\shared_i; \shared_j \bigr]  +
  \mathbf{b}_\text{LA}^{(t)} \bigr) .
\end{split}
\end{equation}
The \pred\ and \unlabeledarc\ versions are analogous.
The output representations $\{ \bm{\psi}^{(t)} \}$ remain task-specific,
and the score is still the inner product between the input representation and
the output representation.

The second variant, which we call \textsc{shared}, uses \emph{only} the shared
encoder $\shared$, and doesn't use task-specific encoders $\{\bilstm^{(t)}\}$.
It can be understood as a special case of \textsc{freda} where the dimensions
of the task-specific encoders are 0.

\subsection{Multitask SDP with Cross-Task Structures}
\label{subsec:multi:ct}

In syntactic parsing, higher-order structures have commonly been used to
model interactions between multiple adjacent arcs in the same dependency tree
\interalia{carreras2007experiments,smith2008dependency,martins2009concise,zhang_greed_2014}.
\citet{lluis2013joint}, in contrast, used second-order structures to
jointly model syntactic dependencies and semantic roles.
Similarly, we use higher-order structures \emph{across} tasks instead of
\emph{within} tasks. 
In this work, we look at interactions between arcs that share the same head
and modifier.\footnote{In the future we hope to model structures over larger motifs, both across
and within tasks, to potentially capture when an arc in one formalism corresponds to a
path in another formalism, for example.}
See Figures~\ref{subfig:2nd_order_factor} and \ref{subfig:3rd_order_factor} for
examples of higher-order cross-task structures.

\paragraph{Higher-order structure scoring.}

Borrowing from \citet{lei2014lowrank}, we introduce a low-rank tensor scoring
strategy that, given a higher-order structure $p$,
models interactions between the first-order structures (i.e., arcs) $p$ is made up of.
This approach builds on and extends the parameter sharing techniques in
\S\ref{subsec:multi:sharing}.
It can either follow \textsc{freda} or \textsc{shared} 
to get the input representations for first-order structures.

We first introduce basic tensor notation.
The \term{order} of a tensor is the number of its dimensions.
The \term{outer product} of two vectors forms a second-order tensor (matrix)
where $\left[\mathbf{u} \otimes \mathbf{v}\right]_{i,j} = u_iv_j$.
We denote the \term{inner product} of two tensors of the same dimensions by
$\left\langle \cdot, \cdot \right\rangle$, which first takes their element-wise
product, then sums all the elements in the resulting tensor.

For example, let $p$ be a labeled third-order structure, including one labeled arc from
each of the three different tasks: $p = \{ p^{(t)}\}_{t\in\mathcal{T}}$.
Intuitively, $s(p)$ should capture every pairwise interaction between the
three input and three output representations of $p$.
Formally, we want the score function to include a parameter for each term in
the outer product of the representation vectors:
$s(p) = $
\begin{align}
\label{score:tensor}
\left\langle
  \bm{\mathscr{W}},
  \bigotimes_{t\in\mathcal{T}}
  \left(\bm{\phi}^{(t)}\left(p^{(t)}\right) \otimes
      \bm{\psi}^{(t)}\left(p^{(t)}\right) \right)
\right\rangle,
\end{align}
where $\bm{\mathscr{W}}$ is a sixth-order tensor of parameters.\footnote{%
This is, of course, not the only way to model interactions
between several representations.
For instance, one could concatenate them and feed them into another MLP.
Our preliminary experiments in this direction suggested that it may be less
effective given a similar number of parameters, but we did not run full
experiments.
}

With typical dimensions of representation vectors, this  leads to an
unreasonably large number of parameters.
Following \citet{lei2014lowrank}, we upper-bound the rank of $\bm{\mathscr{W}}$ by $r$
to limit the number of parameters
($r$ is a hyperparameter, decided empirically).
Using the fact that a tensor of rank at most $r$ can be decomposed into a sum of
$r$ rank-1 tensors \citep{hitchcock1927expression}, we reparameterize $\bm{\mathscr{W}}$ to
enforce the low-rank constraint by construction:
\begin{equation}
\label{eq:tensor_decompose}
\bm{\mathscr{W}} = \sum_{j=1}^{r} {\bigotimes_{t\in\mathcal{T}} 
{\left(\left[\mathbf{U}_{\mathrm{LA}}^{(t)}\right]_{j,:}
 \otimes \left[\mathbf{V}_{\mathrm{LA}}^{(t)}\right]_{j,:}\right)}}, 
\end{equation}
where $\mathbf{U}_{\mathrm{LA}}^{(t)}, \mathbf{V}_{\mathrm{LA}}^{(t)} \in \R^{r\times d}$ are now our
parameters.
$[\cdot]_{j,:}$ denotes the $j$th row of a matrix.
Substituting this back into Equation \ref{score:tensor} and rearranging, the
score function $s(p)$ can then be rewritten as:
\begin{equation}
\label{eq:multiscore_decomposed}
\sum_{j=1}^{r} \prod_{t\in\mathcal{T}}
\left[\mathbf{U}_{\mathrm{LA}}^{(t)}\bm{\phi}^{(t)}\hspace{-.05cm}\left(p^{(t)}\right)\right]_{j}
\left[\mathbf{V}_{\mathrm{LA}}^{(t)}\bm{\psi}^{(t)}\hspace{-.05cm}\left(p^{(t)}\right)\right]_{j}.
\end{equation}
We refer readers to \citet{kolda2009tensor} for mathematical details.

For labeled higher-order structures our parameters consist of the set of six matrices,
$\{\mathbf{U}_{\mathrm{LA}}^{(t)}\} \cup \{\mathbf{V}_{\mathrm{LA}}^{(t)}\}$.
These parameters are shared between second-order and third-order labeled structures.
Labeled \emph{second-order} structures are scored as Equation~\ref{eq:multiscore_decomposed},
but with the product extending over only the two relevant tasks.
Concretely, only four of the representation functions are used rather than all six,
along with the four corresponding matrices from
$\{\mathbf{U}_{\mathrm{LA}}^{(t)}\} \cup \{\mathbf{V}_{\mathrm{LA}}^{(t)}\}$.
\emph{Unlabeled} cross-task structures are scored analogously,
reusing the same representations, but with a separate set of parameter matrices
$\{\mathbf{U}_{\mathrm{UA}}^{(t)}\} \cup \{\mathbf{V}_{\mathrm{UA}}^{(t)}\}$.

Note that we are not doing tensor factorization; we are learning
$\mathbf{U}_{\mathrm{LA}}^{(t)}, \mathbf{V}_{\mathrm{LA}}^{(t)},
\mathbf{U}_{\mathrm{UA}}^{(t)},$ and $\mathbf{V}_{\mathrm{UA}}^{(t)}$
directly, and $\bm{\mathscr{W}}$ is never explicitly instantiated.

\paragraph{Inference and learning.}

Given a sentence, we use \adcubed\ to jointly decode all three formalisms.\footnote{%
Joint inference comes at a cost;
our third-order model is able to decode roughly 5.2 sentences 
(i.e., 15.5 task-specific graphs) per second on a single Xeon E5-2690 2.60GHz CPU.
}
The training objective used for learning is the sum of the losses for individual
tasks.

\subsection{Implementation Details}
\label{sec:implementation}
Each input token is mapped to a concatenation of three real vectors: a 
pre-trained word vector; a randomly-initialized word vector; and a 
randomly-initialized POS tag vector.\footnote{%
There are minor differences in the part-of-speech data provided with the three formalisms. 
For the basic models, we use the POS tags provided with the respective dataset;
for the multitask models, we use the (automatic) POS tags provided with DM. 
}
All three are updated during training. 
We use 100-dimensional \texttt{GloVe} \citep{pennington2014glove} vectors trained over Wikipedia and Gigaword as pre-trained word embeddings. 
To deal with out-of-vocabulary words, we apply word dropout \citep{iyyer2015deep} and randomly replace a word $w$ with a special unk-symbol with probability $\frac{\alpha}{1 + \#(w)}$, where $\#(w)$ is the count of $w$ in the training set. 

Models are trained for up to 30 epochs with Adam \citep{kingma2014adam}, with 
$\beta_1 = \beta_2 = 0.9$, and initial learning rate $\eta_0 = 10^{-3}$. 
The learning rate $\eta$ is annealed at a rate of $0.5$ every 10 epochs \cite{dozat2016deep}. 
We apply early-stopping based on the labeled ${F}_1$ score 
on the development set.\footnote{Micro-averaged labeled ${F}_1$ for the multitask models.}
We set the maximum number of iterations of $\mathrm{AD}^3$ to 500 and round decisions when it doesn't converge. 
We clip the $\ell_2$ norm of gradients to 1 \citep{graves2013generating,sutskever2014sequence}, and we do not use mini-batches. 
Randomly initialized parameters are sampled from a uniform distribution over $\bigl[-\sqrt{6/(d_{\mathrm{r}} + d_{\mathrm{c}})}, \sqrt{6/(d_{\mathrm{r}} + d_{\mathrm{c}})}\bigr]$, where $d_{\mathrm{r}}$ and $d_{\mathrm{c}}$ are the number of the rows and columns in the matrix, respectively. 
An $\ell_2$ penalty of $\lambda = 10^{-6}$ is applied to all weights. 
Other hyperparameters are summarized in Table \ref{tb:hyperparameters}. 

We use the same pruner as \citet{martins2014sdp}, where a first-order feature-rich unlabeled pruning model is trained for each task, and arcs with posterior probability below $10^{-4}$ are discarded. 
We further prune labeled structures that appear less than 30 times in the training set.
In the development set, about 10\% of the arcs remain after pruning, with a recall of around 99\%. 

\begin{table}
\small
\ra{1.1}
\begin{tabulary}{.48\textwidth}{@{}l R@{}}
  \toprule 
  \textbf{Hyperparameter} & \textbf{Value}\\
  \midrule 
  Pre-trained word embedding dimension & 100 \\
  Randomly-initialized word embedding dimension & 25 \\
  POS tag embedding dimension & 25 \\
  Dimensions of representations $\bm{\phi}$
  and $\bm{\psi}$ & 100\\
  MLP layers & 2\\
  BiLSTM layers & 2\\
  BiLSTM dimensions & 200 \\
  Rank of tensor $r$ & 100\\
  $\alpha$ for word dropout & 0.25\\
  \bottomrule 
\end{tabulary}
\caption{Hyperparameters used in the experiments.}
\vspace{-.6cm}
\label{tb:hyperparameters}
\end{table}

\subsection{Experiments}
\label{subsec:multi:experiment}

\paragraph{Experimental settings.}
We compare four multitask variants to the basic model, as well as the two baseline systems introduced in \S\ref{subsec:mono:experiment}.
\begin{compactitem}
  \item
    \textsc{shared1} is a first-order model.
    It uses a single shared BiLSTM encoder, and keeps the inference separate for each task.
  \item
    \textsc{freda1} is a first-order model based on ``frustratingly
    easy'' parameter sharing.
    It uses a shared encoder as well as task-specific ones.
    The inference is kept separate for each task.
  \item
    \textsc{shared3} is a third-order model.
    It follows \textsc{shared1} and uses a single shared BiLSTM encoder,
    but additionally employs cross-task structures and inference.
  \item
    \textsc{freda3} is also a third-order model.
    It combines \textsc{freda1} and \textsc{shared3} by using
    both ``frustratingly easy'' parameter sharing and cross-task structures and inference.
\end{compactitem}
In addition, we also examine the effects of syntax by comparing our models to
the state-of-the-art open track system
\citep{almeida2015sdp}.\footnote{\citet{kanerva2015sdp} was the winner of the gold track, which overall saw
higher performance than the closed and open tracks.
Since gold-standard syntactic analyses are not available in most realistic
scenarios, we do not include it in this comparison.
} 

\begin{table}[tb]
   \begin{subtable}[tb]{\columnwidth}
   \center
    \begin{tabulary}{\columnwidth}{@{}l  rrr  r@{}}
			\toprule

			& \textbf{DM}
			& \textbf{PAS}
			& \textbf{PSD}
			& \textbf{Avg.}\\

			\midrule

			Du et al., 2015
			& 89.1 &  91.3
			& 75.7 & 86.3 \\

			A\&M, 2015 (closed)
			& 88.2 &  90.9
			& 76.4 & 86.0 \\

			A\&M, 2015 (open)$^\dagger$
			& 89.4 &  91.7
			& 77.6 & 87.1 \\

			\textsc{basic}
			& 89.4 &  \ul{92.2}
			& 77.6 & 87.4 \\

			\midrule[\cmidrulewidth]

			\textsc{shared1}
			& 89.7 &  91.9
			& 77.8 & 87.4 \\

			\textsc{freda1}
			& \ul{90.0} &  \ul{92.3}
			& \ul{78.1} & \ul{87.7}\\

			\midrule[\cmidrulewidth]

			\textsc{shared3}
			& \ul{90.3} &  \ul{92.5}
			& \ul{\textbf{78.5}} & \ul{\textbf{88.0}}\\

			\textsc{freda3}
			& \ul{\textbf{90.4}} & \ul{\textbf{92.7}}
			& \ul{\textbf{78.5}} & \ul{\textbf{88.0}}\\


			\bottomrule
		\end{tabulary}
    \caption{Labeled ${F}_1$ score on the in-domain test set.}
    \vspace{.5cm}
    \label{tab:res_id_lf}
  \end{subtable}
  \hfill
  \begin{subtable}[tb]{\columnwidth}
  \center
   \begin{tabulary}{\columnwidth}{@{}l rrr r@{}}
			\toprule

			& \textbf{DM}
			& \textbf{PAS}
			& \textbf{PSD}
			& \textbf{Avg.}\\

			\midrule

			Du et al., 2015
			& 81.8 &  87.2
			& 73.3 & 81.7 \\
			
			A\&M, 2015 (closed)
			& 81.8 &  86.9
			& 74.8 & 82.0 \\
			
			A\&M, 2015 (open)$^\dagger$
			& 83.8 &  87.6
			& 76.2 & 83.3 \\


			\textsc{basic}
			& \ul{84.5} &  \ul{88.3}
			& 75.3 & 83.6 \\

			\midrule[\cmidrulewidth]

			\textsc{shared1}
			& \ul{84.4} &  \ul{88.1}
			& 75.4 & 83.5 \\
			
			\textsc{freda1}
			& \ul{84.9} &  \ul{88.3}
			& 75.8 & \ul{83.9}\\

			\midrule[\cmidrulewidth]

			\textsc{shared3}
			& \ul{\textbf{85.3}} &  \ul{88.4}
			& 76.1 & \ul{84.1}\\

			\textsc{freda3}
			& \ul{\textbf{85.3}} & \ul{\textbf{89.0}}
			& \textbf{76.4} & \ul{\textbf{84.4}} \\
			
			\bottomrule
		\end{tabulary}
    \caption{Labeled ${F}_1$ score on the out-of-domain test set.}
    \label{tab:res_ood_lf}
  \end{subtable}
  \caption{The last columns show the micro-average over the three tasks. $\dagger$ denotes the use of syntactic parses. Bold font indicates best performance among all systems, and underlines indicate statistical significance with Bonferroni correction compared to A\&M, 2015 (open), the strongest baseline system.}
  \vspace{-.3cm}
  \label{tbl:main}
\end{table}

\paragraph{Main results overview.} 
Table~\ref{tab:res_id_lf} compares our models to the best published results
(labeled ${F}_1$ score) on SemEval 2015 Task 18 in-domain test set.
Our basic model improves over all closed track entries in all
formalisms. 
It is even with the best open track system for DM and PSD, but improves on PAS
and on average, without making use of any syntax.
Three of our four multitask variants further improve over our basic
model; \textsc{shared1}'s differences are statistically insignificant.
Our best models (\textsc{shared3}, \textsc{freda3}) outperform the previous
state-of-the-art closed track system by 1.7\% absolute ${F}_1$, and the
best open track system by 0.9\%,  without the use of syntax.

We observe similar trends on the out-of-domain test set (Table~\ref{tab:res_ood_lf}), with the
exception that, on PSD,
our best-performing model's 
improvement over the open-track system of 
\citet{almeida2015sdp} is not statistically significant.

The extent to which we might benefit from syntactic information remains unclear.
With automatically generated syntactic parses, \citet{almeida2015sdp} manage to
obtain more than 1\% absolute improvements over their closed track entry,
which is consistent with the extensive evaluation by
\citet{zhang2016transition}, but
we leave the incorporation of syntactic trees to future work. 
Syntactic parsing could be treated as yet another output
task, as explored in \citet{lluis2013joint} and in the transition-based
frameworks of \citet{henderson-13} and \citet{swayamdipta2016greedy}.

\paragraph{Effects of structural overlap.}
We hypothesized that the overlap between formalisms would enable multitask
learning to be effective;
in this section we investigate in more detail how structural overlap affected
performance.
By looking at \emph{undirected} overlap between unlabeled arcs, we discover
that modeling only arcs in the same direction may have been a design mistake.

DM and PAS are more structurally similar to each other than either
is to PSD.
Table \ref{tab:structural_similarity} compares the structural
similarities between the three formalisms in unlabeled ${F}_1$ score (each
formalism's gold-standard unlabeled graph is used as a prediction of each other
formalism's gold-standard unlabeled graph).
All three formalisms have more than 50\% overlap when ignoring arcs' directions,
but considering direction, PSD is clearly different; PSD reverses the
direction about half of the time it shares an edge with another formalism.
A concrete example can be found in Figure~\ref{fig:formalisms}, where 
DM and PAS both have an arc from \textit{``Last''} to \textit{``week,''} 
while PSD has an arc from \textit{``week''} to \textit{``Last.''}


We can compare \textsc{freda3} to \textsc{freda1} to isolate the effect of
modeling higher-order structures.
Table~\ref{tab:res_id_ulf} shows performance on the development data in both
unlabeled and labeled ${F}_1$.
We can see that \textsc{freda3}'s unlabeled performance improves on DM and PAS,
but \emph{degrades} on PSD.
This supports our hypothesis, and suggests that in future work, a more careful
selection of structures to model might lead to further improvements.

\begin{table}[tb]
\small
\center
\ra{1.2}
		\begin{tabulary}{0.47\textwidth}{@{}l  RRR c RRR@{}}
			\toprule

			& \multicolumn{3}{c}{Undirected}
			& \phantom{}
			& \multicolumn{3}{c}{Directed}\\
			\cmidrule{2-4}
			\cmidrule{6-8}
			& \bf{DM} & \bf{PAS} & \bf{PSD} &
			& \bf{DM} & \bf{PAS} & \bf{PSD}\\

			\midrule
			
			\bf{DM}
			& - & 67.2 & 56.8 &
			& - &  64.2 & 26.1\\
			
			\bf{PAS}
			& 70.0 & - & 54.9 &
			& 66.9 &  - & 26.1\\
			
			\bf{PSD}
			& 57.4 & 56.3 & - &
			& 26.4 &  29.6 & -\\

			\bottomrule
		\end{tabulary}
	\caption{Pairwise structural similarities between the three
          formalisms in unlabeled ${F}_1$ score. Scores from \citet{oepen2015sdp}.} 
           \vspace{-.25cm}
	\label{tab:structural_similarity}
\end{table}

\begin{table}[tb]
\small
\center
		\begin{tabulary}{0.47\textwidth}{@{}l  RR c RR c RR@{}}
			\toprule

			& \multicolumn{2}{c}{\textbf{DM}}
			& \phantom{}
			& \multicolumn{2}{c}{\textbf{PAS}}
			& \phantom{}
			& \multicolumn{2}{c}{\textbf{PSD}}\\

			\cmidrule{2-3}
			\cmidrule{5-6}
			\cmidrule{8-9}
			& U$F$ & L$F$ &
			& U$F$ & L$F$ &
			& U$F$ & L$F$\\

			\midrule


			\textsc{freda1}
			& 91.7 &  90.4 &
			& 93.1 & 91.6 &
			& 89.0 & 79.8\\


			\textsc{freda3}
			& 91.9 & 90.8 &
			& 93.4 & 92.0 &
			& 88.6 & 80.4 \\


			\bottomrule
		\end{tabulary}
	\caption{Unlabeled (U$F$) and labeled (L$F$) parsing
          performance of \textsc{freda1} and \textsc{freda3} on the development set of SemEval  2015 Task 18.
          \vspace{-.25cm}
	\label{tab:res_id_ulf}}
\end{table}

\section{Related Work}
\label{sec:related}
We note two important strands of related work.

\paragraph{Graph-based parsing.}
Graph-based parsing was originally invented to handle non-projective syntax
\interalia{mcdonald2005online,koo2010dual,martins2013turboparser}, but has been
adapted to semantic parsing
\interalia{flanigan2014amr,martins2014sdp,thomson2014sdp,kuhlmann2014sdp}.
Local structure scoring was traditionally done with linear models
over hand-engineered features, but lately, various forms of representation
learning have been explored to learn feature combinations
\interalia{lei2014lowrank,taubtabib2015template,pei2015effective}.
Our work is perhaps closest to those who used BiLSTMs to encode inputs
\citep{kiperwasser2016simple,kuncoro-16,wang2016graph,dozat2016deep,ma2017neural}.

\paragraph{Multitask learning in NLP.}
There have been many efforts in NLP to use joint learning to replace pipelines,
motivated by concerns about cascading errors. 
\citet{collober2008unified} proposed sharing the same word
representation while solving multiple NLP tasks.
\citet{zhang2016stack} use a continuous stacking model 
for POS tagging and parsing. 
\citet{ammar2016many} and \citet{guo2016universal} explored parameter sharing
for multilingual parsing.
\citet{johansson2013training} and \citet{kshirsagar2015frame} applied
ideas from domain adaptation to multitask learning.
 Successes in multitask learning have been enabled by advances in representation
 learning as well as earlier explorations of parameter sharing
 \citep{ando2005framework,blitzer-06,daume2007frustratingly}.

\section{Conclusion}
\label{sec:conclusion}

We showed two orthogonal ways to apply deep multitask learning to
graph-based parsing.
The first shares parameters when encoding tokens in the input with recurrent neural
networks, and the second introduces interactions between output
structures across formalisms.
Without using syntactic parsing, these approaches outperform even
state-of-the-art semantic dependency parsing systems that use syntax.
Because our techniques apply to labeled directed graphs in general, they
can easily be extended to incorporate more formalisms, semantic or otherwise.
In future work we hope to explore cross-task scoring and inference for tasks where parallel
annotations are not available.
Our code is open-source and available at
\url{https://github.com/Noahs-ARK/NeurboParser}\stcomment{Not necessary?}.

\section*{Acknowledgements}
We thank the Ark, Maxwell Forbes, Luheng He, Kenton Lee, Julian Michael, and
Jin-ge Yao for their helpful comments on an earlier version of this draft,
and the anonymous reviewers
for their valuable feedback.
This work was supported by
NSF grant IIS-1562364
and DARPA grant FA8750-12-2-0342 funded under the DEFT program.
\bibliography{ml_sdp}
\bibliographystyle{acl_natbib}

\end{document}